\documentclass{article}

\usepackage{microtype}
\usepackage{graphicx}
\usepackage{subfigure}
\usepackage{booktabs} 

\usepackage{hyperref}

\usepackage[accepted]{icml2025} 

\usepackage{amsmath}
\usepackage{amssymb}
\usepackage{mathtools}
\usepackage{amsthm}

\usepackage[capitalize,noabbrev]{cleveref}

\theoremstyle{plain}
\newtheorem{theorem}{Theorem}[section]

\newtheorem{corollary}[theorem]{Corollary}
\theoremstyle{definition}

\theoremstyle{remark}

\usepackage[textsize=tiny]{todonotes}

\icmltitlerunning{Do Graph Diffusion Models Accurately Capture and Generate Substructure Distributions?}

\begin{document}

\twocolumn[
\icmltitle{Do Graph Diffusion Models Accurately Capture and Generate Substructure Distributions?}



\icmlsetsymbol{equal}{*}

\begin{icmlauthorlist}
\icmlauthor{Xiyuan Wang}{iai}
\icmlauthor{Yewei Liu}{xk}
\icmlauthor{Lexi Pang}{xk}
\icmlauthor{Siwei Chen}{xk}
\icmlauthor{Muhan Zhang}{iai}
\end{icmlauthorlist}

\icmlaffiliation{xk}{School of Electronics Engineering and Computer Science}
\icmlaffiliation{iai}{Institute for Artificial Intelligence, Peking University}

\icmlcorrespondingauthor{Muhan Zhang}{wangxiyuan@pku.edu.cn, muhan@pku.edu.cn}
\icmlkeywords{Machine Learning, ICML}

\vskip 0.3in
]



\printAffiliationsAndNotice{}  

\begin{abstract}
Diffusion models have gained popularity in graph generation tasks; however, the extent of their expressivity concerning the graph distributions they can learn is not fully understood. Unlike models in other domains, popular backbones for graph diffusion models, such as Graph Transformers, do not possess universal expressivity to accurately model the distribution scores of complex graph data. Our work addresses this limitation by focusing on the frequency of specific substructures as a key characteristic of target graph distributions. When evaluating existing models using this metric, we find that they fail to maintain the distribution of substructure counts observed in the training set when generating new graphs. To address this issue, we establish a theoretical connection between the expressivity of Graph Neural Networks (GNNs) and the overall performance of graph diffusion models, demonstrating that more expressive GNN backbones can better capture complex distribution patterns. By integrating advanced GNNs into the backbone architecture, we achieve significant improvements in substructure generation.
\end{abstract}

\section{Introduction}

Diffusion models have emerged as a powerful approach for graph generation, demonstrating remarkable success across a wide range of applications, including molecular design, synthetic graph creation, and social network modeling. By progressively corrupting a graph with noise and learning to reverse this process, these models generate novel graphs from random noise while preserving essential structural properties. They have outperformed traditional generative models in key evaluation metrics such as uniqueness, novelty, and molecular validity, making them a compelling choice for diverse real-world applications.  

Despite these advances, evaluating the quality of generated graphs remains a fundamental challenge. Traditional evaluation metrics are often designed with specific downstream tasks in mind, rather than directly assessing the statistical similarity between the training and generated graph distributions. This limitation can result in misleadingly optimistic evaluations, where models appear to perform well despite failing to faithfully replicate the underlying distribution. For instance, molecule validity—a widely used metric—can easily saturate, reaching near-perfect scores without accurately capturing the true diversity and structural fidelity of generated molecules.  

To address these limitations, we introduce a novel evaluation metric based on subgraph count distributions. The motivation behind this approach is straightforward: if two sets of graphs originate from different distributions, there must be discrepancies in the frequencies of certain subgraphs. Unlike traditional metrics, which may overlook fine-grained structural details, our method provides a more direct measure of distributional similarity by assessing how well a generative model preserves key structural motifs, such as cycles and functional substructures in molecules. These substructures are often crucial determinants of graph properties, influencing molecular stability, network connectivity, and functional characteristics.  

Our empirical analysis reveals a critical gap: existing diffusion models fail to accurately preserve the subgraph frequency distributions observed in the training data. This issue persists in both synthetic datasets with predefined subgraph patterns and real-world datasets, where structural integrity is essential for meaningful applications. The inability of current models to generate graphs with the correct subgraph statistics suggests a fundamental limitation in their design.  

To investigate the root cause of this limitation, we analyze the architectural expressivity of diffusion models, particularly in their ability to estimate the score function—a mathematical representation of the noisy graph distribution at each step of the denoising process. Our theoretical analysis shows that the score function can be decomposed into graph polynomials, whose coefficients are directly influenced by subgraph counts in the training set. As the complexity of these subgraphs increases, so does the complexity of the score function, often surpassing the expressive capacity of standard Graph Neural Networks (GNNs) used as backbones in diffusion models.  

To overcome this challenge, we propose enhancing the expressivity of GNN architectures within diffusion models. By leveraging higher-order GNNs capable of capturing complex graph polynomials, we enable more precise score function estimation. This, in turn, leads to a more accurate reproduction of subgraph structures, thereby improving the overall fidelity of generated graphs.  

In summary, our work contributes to both the evaluation and generation of graph structures by:  
\begin{itemize}
    \item  Introducing a robust metric based on subgraph count distributions, which provides a more faithful assessment of generative model performance.  \item Identifying a fundamental expressivity limitation in current graph diffusion models that hinders their ability to capture complex substructures. 
    \item Proposing more expressive GNN architectures to enhance score function estimation and improve subgraph preservation in generated graphs. 
\end{itemize}
By bridging the gap between theory and practice, our findings offer valuable insights for improving the structural integrity of generated graphs, ensuring that they not only resemble the training data statistically but also retain their functional significance in downstream applications.

\section{Preliminary}
For a matrix $Z \in \mathbb{R}^{a \times b}$, we denote by $Z_i \in \mathbb{R}^b$ the $i$-th row of $Z$, treated as a column vector, and by $Z_{ij} \in \mathbb{R}$ its element located at the intersection of the $i$-th row and $j$-th column.

A \textit{graph} is represented as $\mathcal{G} = (V, E, X)$, where $V = \{1, 2, \ldots, n\}$ is the set of nodes, $E \subseteq V \times V$ is the set of edges connecting pairs of nodes, and $X \in \mathbb{R}^{n \times d}$ is the node feature matrix. Each row in $X$, denoted $X_v$, represents the feature vector associated with node $v$. The edge set $E$ can also be expressed through an adjacency matrix $A \in \{0, 1\}^{n \times n}$. In this matrix, $A_{uv}$ equals 1 if there exists an edge between nodes $u$ and $v$, and 0 otherwise.

A \textit{subgraph} $G^{S} = (V^{{S}}, E^{S}, X)$ is defined by a subset of nodes $V^{S} \subseteq V$ and a subset of edges $E^{S} \subseteq E$. If the edge set is induced by the node subset, e.g. $E^{S}=\{(i, j)|(i, j)\in E, i \in V^{S}, j\in V^{S}\}$, we say it is an induced subgraph.

Two graphs are considered \textit{isomorphic} if one can be transformed into the other via a permutation of the node indices. The \textit{subgraph-count} $C_S(G)$ represents the number of subgraphs of $G$ that are isomorphic to the pattern $S$. When considering a distribution $p$ over graphs and a subgraph pattern $S$, $p^{(S)}$ denotes \textbf{the distribution of counts of the pattern $S$ across natural numbers}. 

\section{Analyzing Existing Graph Generation Models' Capacity to Generate Substructures}
Despite extensive research in graph generation, evaluating the quality of generated samples remains a challenge. Traditional evaluation metrics are often designed for specific downstream tasks rather than for directly comparing the statistical distribution of the training set and generated samples. This misalignment can lead to overly optimistic evaluations, even when the generated distributions significantly deviate from the original data. For example, metrics like molecular validity can quickly saturate, achieving high scores without accurately reflecting the diversity or structural fidelity of the generated molecules. To address this limitation, we propose measuring the distance between subgraph distributions in the generated samples and the training dataset as a more reliable indicator of generation quality. Our analysis of existing models shows that most perform poorly in preserving substructure distributions, underscoring a key weakness in current approaches.

\subsection{Distance between Substructure Distribution As Metric}

While previous studies have not explicitly evaluated generation quality through subgraph counts, many existing metrics can be framed in terms of subgraph distributions. For instance, \textit{novelty}, which measures the ratio of generated molecules that do not appear in the training set, can be interpreted using subgraph counts. Specifically, if a subgraph pattern appears in the training set with a nonzero count and a generated molecule contains the same number of nodes and edges as that pattern, then the generated molecule is considered part of the training set. Otherwise, it is novel.

Similarly, the \textit{NSPDK kernel} represents molecules as vectors of subgraph counts and computes distances between these vectors to quantify differences between graphs—the smaller the distance, the better. This highlights that measuring the distance between substructure distributions can unify various existing evaluation metrics. 

Moreover, substructure counts provide interpretable and meaningful insights for downstream tasks. For example, cycle structures play a crucial role in determining molecular properties and stability, and generating cycle correctly is crucial for molecule generative models. Additionally, subgraph counts offer a comprehensive perspective—for any two distinct graph distributions, there must be some subgraph patterns where their distributions differ, and a model may perform well on generating some subgraphs.

To quantify these differences, given a subgraph pattern \( S \), a training set graph distribution \( p \), and a generated graph distribution \( q \), we use the total variation distance (TV) between \( p^{(S)} \) and \( q^{(S)} \):

\begin{equation}
    TV = \frac{1}{2}\sum_{i=0,1,...}|p^{(S)}(i)-q^{(S)}(i)|
\end{equation}

The TV distance ranges from 0 to 1, where a lower value indicates a smaller discrepancy between the training set and generated samples, implying a stronger ability of the model to replicate subgraph distributions.

\subsection{Evaluation of Existing Graph Generation Models}

We evaluate models under two settings:
\begin{itemize}
    \item \textbf{Subgraph count preservation} – The training set consists of graphs that each contain only a single subgraph type to focus on the expressivity of model, and we test whether the model can preserve that subgraph in generated graphs.
    \item \textbf{Subgraph distribution preservation} – The training set consists of real-world graphs containing varied subgraph counts, and we assess how well the generated graphs replicate this distribution.
\end{itemize}

\begin{table*}[t]
    \centering
    \caption{Evaluation of existing models on synthetic datasets (top) and QM9 (bottom). The distance is estimated using 100 generated samples for synthetic datasets and 1,000 for QM9. Columns represent different subgraph structures. Blank cells indicate omitted results due to data rarity or implementation issues.  Subgraphs includes cycles of length 3 to 8 (c3-c8), a 3-cycle sharing an edge with a 4-cycle (c3c4), and line structures (l5-l7). Due to rarity, some subgraphs are omitted in QM9. Additionally, we exclude HGGT results on QM9 due to issues with its official implementation. }\label{tab:syncount}
    \vskip 0.1 in
    \setlength{\tabcolsep}{3pt}
    \begin{tabular}{lccccccccccccc}
    \toprule
    Model & c3 & c4 & c5 & c6 & c7 & c8 & c3c4 & c5c5 & c5c6 & c6c6 & l5 & l6& l7 \\
    \midrule
    GDSS~\citep{ScorebasedGraphDSS}&   0.12 & 0.33 & 0.97 & 1.00 & 1.00 & 0.95 & 1.00 & 1.00 & 1.00 & 1.00 & 1.00 & 0.99 & 0.99\\
    DiGress~\citep{DiGressDDGraph}& 0.00& 0.00 & 0.00 & 0.02& 0.04 & 0.15 & 0.03 & 0.36 & 0.39& 0.73 & 0.08 & 0.15 & 0.28\\
    HGGT~\citep{HGGT} & 0.00 &0.00 & 0.94& 0.91& 0.88& 0.61& 0.57& 0.0& 0.84 & 1.00 & 1.00 & 0.85 & 1.00\\
    Grum~\citep{Grum} & 0.00 & 0.00 & 0.00 & 0.00 & 0.00 & 0.00 & 0.01 & 0.12 & 0.06 & 0.12 & 0.00 & 0.00 & 0.00 \\
    \midrule
    GDSS~\citep{ScorebasedGraphDSS}& 0.076 & 0.036 & 0.141 & 0.020 & 0.019 & 0.006 & 0.065 & 0.019 & 0.004 & - & 0.090 & 0.038 & 0.010\\
    DiGress~\citep{DiGressDDGraph}& 0.042 & 0.045 & 0.023 & 0.002 & 0.017 & 0.006 & 0.029 & 0.011 & 0.000 & - & 0.041 & 0.039 & 0.011\\
    Grum~\citep{Grum} & 0.015 & 0.077 & 0.033 & 0.008 & 0.013 & 0.002 & 0.007 & 0.011 & 0.003 & - & 0.026 & 0.023 & 0.002\\
    \bottomrule
    \end{tabular}
\end{table*}

In the subgraph count preservation setting (top half of Table~\ref{tab:syncount}), each column represents a model’s ability to preserve specific subgraph structures. We trained separate models for each subgraph type, ensuring the training set contained only graphs with a single instance of that subgraph. The TV distance here directly corresponds to the proportion of generated samples that fail to preserve the expected subgraph count. 

Most models perform well on simple subgraphs like 3-cycles, but surprisingly, some seemingly simple structures, such as 7-line, are almost entirely unlearnable—almost no generated graphs retain these structures. A notable exception is Grum, which perfectly preserves all subgraphs because it directly copies graphs from the training set as templates. However, this comes at the cost of low novelty, as most generated graphs are nearly identical to training data.

For the subgraph distribution preservation setting (bottom half of Table~\ref{tab:syncount}), we evaluate models on QM9, a widely used molecular dataset. Since real-world datasets contain graphs with varied subgraph counts, the distributions are broader than in synthetic datasets, resulting in lower TV distances. However, models that perform well on synthetic datasets also perform well on QM9. 

Notably, while Grum again achieves low TV distances, its reliance on training set templates leads to a major drawback: low novelty. Only about 20\% of its generated molecules are distinct from the training set, highlighting its limited generalization ability.

\section{How Can Graph Diffusion Models Generate Substructures?}

To understand why graph diffusion models struggle to accurately capture the training set distribution, we analyze the problem theoretically. According to Theorem 2 in \citet{DiffusionError}, the total variation distance between the generated distribution and the target distribution in a Denoising Diffusion Probabilistic Model (DDPM) is controlled by the error in score estimation. Thus, we first derive the analytical form of the score function that the model must learn and assess whether common Graph Neural Networks (GNNs) are expressive enough to approximate it.

\subsection{Graph Diffusion Model}

For simplicity, we analyze a basic graph diffusion model where Gaussian noise is gradually added to the adjacency matrix, while node features are ignored.  
\begin{itemize}
    \item At time step \( t=0 \), the adjacency matrix \( A_0 \in \{0,1\}^{n\times n} \) represents a graph sampled from the training distribution \( p_0 \).
    \item Throughout this analysis, we assume all graphs in \( p_0 \) have exactly \( n \) nodes and \( m \) edges. This assumption does not reduce the generality of our results because most GNNs can easily compute the number of nodes and edges, allowing them to learn separate score functions for graphs of different sizes.  
\end{itemize}
At a later time step \( t \), the noisy adjacency matrix \( A_t \) follows a distribution \( p_t \). Given a Gaussian noise process, the transition probability is:
\begin{equation}
p_t(A_t | A_0) = \frac{1}{\beta_t \sqrt{2\pi}} \exp\left(-\frac{1}{2\beta_t^2} \|A_t - \alpha_t A_0\|_F^2 \right),
\end{equation}

where \( \alpha_t \) and \( \beta_t \) are constants dependent on \( t \) and the noise schedule.  

The model's \textit{score function}, which the GNN needs to learn, is defined as:  
\begin{equation}
\nabla \log p_t(A_t),
\end{equation}
which takes the noisy adjacency matrix \( A_t \) as input. 

\subsection{Graph Polynomial Bases}  

To analyze whether GNNs can learn this score function, we use \textit{graph polynomial bases}, which provide a framework for expressing functions over graphs.  

There are two primary approaches for evaluating GNN expressivity:  
\begin{itemize}
    \item Graph Isomorphism Tests: These approaches compare a GNN’s ability to distinguish non-isomorphic graphs, often using Weisfeiler-Lehman (WL) tests \citep{GIN, WLgoNeuron, WLgoSparse, klWL}.
    \item Function Approximation with Polynomial Bases: This approach studies whether GNNs can approximate arbitrary graph functions by decomposing them into basis functions \citep{IGN, GPoly}.  
\end{itemize}
Since our goal is to approximate a specific function (the score function) rather than differentiate graphs, and given that the input graphs contain continuous noise, the second approach is more suitable.  

\citet{GPoly} introduced a set of permutation-equivariant and permutation-invariant polynomial bases that can be used to approximate any continuous graph function with the corresponding symmetry properties.  

Each invariant polynomial basis \( Q_S: \mathbb{R}^{n \times n} \to \mathbb{R} \) corresponds to a graph \( S \) with nodes \( 1, 2, ..., k \) and edge set \( E^{(S)} \), defined as:  

\begin{equation}\label{equ:diffcondi}
Q_{S}(A) = \frac{1}{n!} \sum_{j_1 \neq j_2 \neq ... \neq j_k \in [n]} \prod_{(a,b) \in E^{(S)}} A_{j_a j_b}.
\end{equation}

Each node \( a \) in \( S \) is assigned an index \( j_a \), and each edge \( (a,b) \) contributes the adjacency term \( A_{j_a j_b} \).  

This formulation is directly linked to subgraph counting: if \( S \) is a simple graph and \( A \) contains only binary values (0 or 1), then:

\begin{equation}
n! Q_{S}(A) = |Aut(S)| C_S(A),
\end{equation}

where \( C_S(A) \) is the number of subgraphs in \( A \) that are isomorphic to \( S \), and \( |Aut(S)| \) is the size of \( S \)'s automorphism group.  

Similarly, equivariant polynomial bases \( \tilde{Q}_{(c,d), S}: \mathbb{R}^{n \times n} \to \mathbb{R}^{n \times n} \) extend this concept to functions that depend on specific node pairs:  
\begin{equation}
\tilde Q_{(c,d),S}(A)_{ij} = \frac{1}{n!} \sum_{\substack{j_1 \neq j_2 \neq ... \neq j_k \in [n] \\ j_c = i, j_d = j}} \prod_{(a,b) \in E^{(S)}} A_{j_a j_b}.
\end{equation}
When \( S \) is a simple graph with binary adjacency values:

\begin{equation}
n! \tilde Q_{(c,d),S}(A)_{ij} = |Aut(S)| C_{ij, cd, S}(A),
\end{equation}

where \( C_{ij, cd, S}(A) \) counts the number of subgraphs isomorphic to \( S \) in \( A \), with nodes \( i, j \) mapped to nodes \( c, d \) in \( S \). It can also be interpretted as link-level count subgraphs that rooted in the link.

\subsection{Score Function expressed with Graph Polynomial}

To understand how graph diffusion models generate substructures, we derive an explicit expression for the score function using graph polynomial bases. This allows us to analyze whether the backbone models, typically GNNs, are expressive enough to learn the required function.
\begin{theorem}
With the diffusion process in Equation~\ref{equ:diffcondi}, assuming input graph distribution is permutation invariant and contains only graph with $n$ nodes and $m$ edges, the score function 
\begin{equation}\label{equ:fullscore}
    \nabla \log p_t(A_t) = \frac{1}{\beta_t^2}A_t+\frac{\alpha_t}{\beta_t^2}\frac{1}{G_t(A_t)}F_t(A_t),
\end{equation}
where $F_t(A_t): \mathbb{R}^{n\times n}\to \mathbb{R}^{n\times n}, G_t(A_t): \mathbb{R}^{n\times n}\to \mathbb{R}$ are functions as follows,
\begin{align}\label{equ:scoreF}
F_t(A_t)
& = 
\sum_{k=0}^\infty\sum_{ij\in [n]^2}\sum_{\mathbf{a}\in [n]^{2k}}\frac{\alpha_t^k}{\beta_t^{2k}k!}\notag{}\\ &[\mathbb{E}_{A_0\sim p_0}Q_{S_{ij\mathbf{a}}}(A_0)]\tilde Q_{T_{ij\mathbf{a}}}(A_t),
\end{align}
\begin{align}\label{equ:scoreG}
G_t(A_t)
&=
\sum_{k=0}^\infty\sum_{ij\in [n]^2}\sum_{\mathbf{a}\in [n]^{2k}}\frac{\alpha_t^k}{\beta_t^{2k}k!} \notag{}\\&[\mathbb{E}_{A_0\sim p_0}Q_{S_{\mathbf{a}}}(A_0)]Q_{S_{\mathbf{a}}}(A_t),
\end{align}
where 
\begin{itemize}
    \item $\tilde Q, Q$ are equivariant and invariant graph bases, respectively.
    \item $S_{ij\mathbf{a}}$ is a graph formed by nodes $\{i,j\}$ and numbers in $\mathbf{a}$ with edge $(i,j)$ and $\{\!\{(a_{2l-1},a_{2l})|l=1,2,...,k\}\!\}$.
    \item $S_{\mathbf{a}}$ is a graph formed by nodes with id in $\mathbf{a}$ and $\{\!\{(a_{2l-1},a_{2l})|l=1,2,...,k\}\!\}$.
    \item $T_{ij\mathbf{a}}$ is a tuple of indice $(i,j)$ and graph with nodes $\{i,j\}$ and numbers in $\mathbf{a}$ with edge $\{\!\{(a_{2l-1},a_{2l})|l=1,2,...,k\}\!\}$.
\end{itemize}
\end{theorem}

This theorem shows that the score function of a diffusion model consists of two main terms:  
\begin{itemize}
    \item A Linear Component: \( \frac{1}{\beta_t^2} A_t \), which directly depends on the noisy adjacency matrix. This term can be learned relatively easily by most GNNs.
    \item A Nonlinear Component: \( \frac{\alpha_t}{\beta_t^2} \frac{1}{G_t(A_t)} F_t(A_t) \), which involves graph polynomial basis functions that encode structural information from the training set.  
\end{itemize}
The complexity of the second term depends on whether a GNN can express the polynomial basis functions \( Q_{S_{\mathbf{a}}}(A_t) \) and \( \tilde Q_{T_{ij\mathbf{a}}}(A_t) \). These functions capture the presence and frequency of substructures, meaning that the expressivity of the GNN determines how well it can reconstruct the true score function.

Most GNN architectures can naturally learn simple transformations like linear combinations, divisions, and identity mappings of \( A_t \). However, for accurate diffusion modeling, the GNN must also learn the more complex polynomial terms in \( F_t(A_t) \) and \( G_t(A_t) \).  

The coefficients in these terms involve expected subgraph counts in the training set, represented as \( \mathbb{E}_{A_0\sim p_0}Q_{S_{\mathbf{a}}}(A_0) \) and \( \mathbb{E}_{A_0\sim p_0}Q_{S_{ij\mathbf{a}}}(A_0) \). These values reflect how frequently certain subgraphs appear in the training data.  
\begin{itemize}
    \item If a specific subgraph never appears in the training set, its corresponding basis function will have a zero coefficient, simplifying the score function.  
    \item Conversely, if the training set contains diverse subgraphs, the score function will require the model to approximate many polynomial bases, demanding higher GNN expressivity.  
\end{itemize}

This insight leads to the following corollary:  
\begin{corollary}
Given $\mathcal{S}$, the set of all subgraphs exists in the training set, let $\mathcal{S}'$ denote the set of all subgraphs selecting marking two special nodes and adding one edge between them from subgraphs in $\mathcal{S}$. If a model can express all graph polynomial basis $Q_{s}$ for $s\in \mathcal{S}$ and $T_{s'}$ for all $s'\in \mathcal{S}'$, then this model can express extract score function on this dataset.
\end{corollary}
Therefore, when the backbone can count all subgraphs in the training set and link-level count all subgraphs rooted in some link, then it can express the score function. 

\subsection{Graph Diffusion Model with Expressive Backbone}

\begin{table*}[t]
    \centering
    \caption{TV score of GDSS with expressive backbones on synthetic datasets. }\label{tab:syncount2}
    \vskip 0.1 in
    \setlength{\tabcolsep}{3pt}
    \begin{tabular}{lccccccccccccc}
    \toprule
    Model & c3 & c4 & c5 & c6 & c7 & c8 & c3c4 & c5c5 & c5c6 & c6c6 & l5 & l6& l7 \\
    \midrule
    GDSS&   0.12 & 0.33 & 0.97 & 1.00 & 1.00 & 0.95 & 1.00 & 1.00 & 1.00 & 1.00 & 1.00 & 0.99 & 0.99\\
    PPGN & 0.00 & 0.00 & 0.01 & 0.02 & 0.24 & 0.19 & 0.58 & 0.94 & 0.88 & 0.98 & 0.52 & 0.58 & 0.65\\
    NGNN & 0.04 & 0.10 & 0.00 & 0.37 & 0.64 & 0.88 & 0.59 & 0.97 & 0.98 & 0.98 & 0.90 & 0.66 & 0.82\\
    SSWL & 0.00 & 0.00 & 0.00 & 0.35 & 0.78 & 0.83 & 0.74 & 1.00 & 0.95 & 0.97 & 0.92 & 0.81 & 1.00\\
    \bottomrule
    \end{tabular}
\end{table*}

Though we do not prove that whether models with lower expressivity can express the score function, in experiments we found this expressivity bound is meaningful. Our experiments confirm that graph diffusion model with more expressive backbones can approximate the score function better. The backbone we use includes:
\begin{itemize}
    \item PPGN~\citep{PPGN} can count cycles up to length 7 but only performs link-level counting for cycles up to length 6.  
    \item NGNN~\citep{NGNN} and SSWL~\citep{SSWL} can only count cycles up to length 6 and perform link-level counting for cycles up to length 5. 
\end{itemize}
The results are shown in Table~\ref{tab:syncount2}. In general, PPGN $>$ SSWL $>$ NGNN in performance, which aligns well with the expressivity order. Moreover, their different expressivity limits directly affect generation quality:  
\begin{itemize}
    \item PPGN can accurately generate cycles up to length 6, but struggles beyond that.  
    \item NGNN and SSWL can accurately generate cycles up to length 5, failing for larger structures. 
\end{itemize}
This empirical evidence supports our theoretical claim: a model’s ability to count and track subgraphs directly determines how well it can generate complex structures in graph diffusion models.

To conclude, we have shown that the score function of a graph diffusion model can be expressed in terms of graph polynomial bases, with coefficients tied to subgraph counts in the training set. This formulation reveals that:  
\begin{itemize}
    \item Graph diffusion models inherently rely on subgraph structure, making subgraph counting ability crucial for generation quality.  
    \item GNN expressivity determines the ability to approximate the score function, with limited models struggling to learn complex structures.  
    \item Empirical results align with theoretical predictions, confirming that models with limited subgraph-counting ability fail to generate large cycles.  
\end{itemize}
These insights provide a principled framework for evaluating and improving the expressivity of diffusion models for graph generation. Future work could explore architectures explicitly designed to capture high-order substructures, potentially improving generation fidelity and diversity.  

\section{Related Work}
\subsection{Diffusion Models for Graphs}  
Diffusion models have gained significant traction in graph generation tasks. Early approaches, such as \citet{ScorebasedGraph}, introduced score-based methods that applied Gaussian perturbations to continuous adjacency matrices, ensuring permutation invariance in generated graphs. Building on this, \citet{ScorebasedGraphDSS} extended the framework to incorporate both node attributes and edges using Stochastic Differential Equations (SDEs). However, these models relied on continuous Gaussian noise, which is inherently misaligned with the discrete nature of graphs.  

To address this issue, \citet{DiffusionGraphBenefitDiscrete} introduced a discrete diffusion model tailored for unattributed graphs, demonstrating the advantages of discrete noise over continuous perturbations in graph generation. Among the most advanced diffusion-based graph generation models, DiGress~\citep{DiGressDDGraph} employs a discrete diffusion process, where noise is introduced by iteratively modifying edges and altering node categories. \citet{SaGessSamplingGraph} further enhanced DiGress by proposing a divide-and-conquer sampling framework, which improves scalability by generating graphs at the subgraph level. Another notable approach, Latent Graph Diffusion (LGD)~\citep{LGD}, first encodes graphs into a latent space using an autoencoder and then applies continuous noise in this transformed space.  

Despite these advancements, most works focus primarily on designing diffusion noise processes, while the choice of backbone architectures—which determine how well the model captures graph structure—remains largely overlooked. These models predominantly employ graph transformers, yet their expressivity in capturing fine-grained substructure distributions is rarely analyzed in depth.

\subsection{Expressivity of Graph Neural Networks}  
The expressivity of Graph Neural Networks (GNNs)—which defines the range of functions a model can learn—is crucial for capturing complex graph distributions. Traditional GNNs, particularly those based on Message Passing Neural Networks (MPNNs)~\citep{MPNN}, update node representations by aggregating information from their neighbors. However, these architectures struggle to capture higher-order dependencies, limiting their ability to model complex graph structures accurately.  

To overcome these limitations, High-Order GNNs (HOGNNs)~\citep{NGNN, SSWL, PPGN} extend message passing by generating tuple-based representations, enabling richer structural encoding. An alternative perspective on GNN expressivity involves analyzing the graph polynomial bases that a model can approximate~\citep{GPoly}. This perspective aligns with subgraph counting, as graph polynomial functions can effectively encode structural motifs in a graph.  

Enhanced GNN architectures, capable of accurately estimating complex graph polynomials, can significantly improve score function modeling in diffusion models. This is particularly critical for preserving key substructures in generated graphs, ensuring high-quality and structurally consistent outputs. In this work, we take a graph polynomial decomposition approach to analyze the expressivity required for diffusion models to accurately capture substructure distributions. By establishing a direct link between score function estimation and GNN expressivity, we provide insights into how backbone architectures influence the fidelity of generated graphs.

\section{Conclusion}  
In this work, we introduce subgraph count distribution distance as a metric for evaluating the generation quality of graph generative models. Surprisingly, our analysis reveals that existing models struggle to generate even simple structures accurately. To understand this limitation, we investigate the expressivity of the backbone models and find that fine-grained substructure generation requires more expressive GNN architectures. Based on this insight, we propose using high-order GNNs as backbone models to improve the ability of diffusion-based graph generators to capture substructures effectively.

\section{Limitation}
Our theoretical analysis primarily focuses on GDSS-like diffusion processes and does not directly extend to discrete diffusion models, flow-based models, or other types of graph generative approaches. However, as shown in Table~\ref{tab:syncount}, the inability to capture subgraph distributions appears to be a general issue across various generative models. In future work, we aim to extend our theoretical framework to a broader range of generative architectures, ensuring a more comprehensive understanding of substructure learning in graph generation.

\section*{Impact Statement}

This paper presents work whose goal is to advance the field of graph representation learning and will improve the design of graph generation models. There are many potential societal consequences of graph learning improvement, such as accelerating drug discovery, improving neural architecture search. None of them we feel need to be specifically highlighted here for potential risk.


\bibliography{example_paper}
\bibliographystyle{icml2025}

\newpage
\appendix
\onecolumn

\section{Graph Diffusion Models' Score Function}

\begin{equation}
\nabla\log p_t(A_t)=\frac{1}{p_t(A_t)}\mathbb E_{A_0\sim p_0}\nabla  p_t(A_t|A_0)
\end{equation}

Put the detail form of $p_t(A_t|A_0)$ into it leads to

\begin{align}
\nabla p_t(A_t|A_0)
=\frac{1}{\beta_t^2}A_t+\frac{\alpha_t}{\beta_t^2}\frac{\mathbb{E}_{A_0\sim p_0}p_t(A_t|A_0)A_0}{\mathbb{E}_{A_0\sim p_0}p_t(A_t|A_0)}
\end{align}
Can GNN with $A_t, t$ as input express this? As neural network can express constant $\alpha_t, \beta_t$, identity mapping $A_t\to A_t$, addition, multiplication, and division, the problem can be converted to whether GNN can express $\mathbb{E}_{A_0\sim p_0}p_t(A_t|A_0)A_0$, and $\mathbb{E}_{A_0\sim p_0}p_t(A_t|A_0)$. 

Considering $\mathbb{E}_{A_0\sim p_0}p_t(A_t|A_0)A_0$, according to Equation~\ref{equ:diffcondi}, it is equivalent to
\begin{equation}
    \frac{\exp{(-\frac{m\alpha_t^2}{2\beta_t^2})}}{\beta_t\sqrt{2\pi}}\exp{(-\frac{\Vert A_t\Vert_F^2}{2\beta_t^2})}\mathbb{E}_{A_0\sim p_0}[A_0\exp{(\frac{\alpha_t}{\beta_t^2}\langle A_0, A_t\rangle)}],
\end{equation}
where $\frac{\exp{(-\frac{m\alpha_t^2}{2\beta_t^2})}}{\beta_t\sqrt{2\pi}}$ is a constant irrelavant to $A_t$, and $\exp{(-\frac{\Vert A_t\Vert_F^2}{2\beta_t^2})}$ is a function of $\Vert A_t\Vert_F$. Assuming model can compute the norm, then model need to compute 

\begin{equation}
    \mathbb{E}_{A_0\sim p_0}[A_0\exp{(\frac{\alpha_t}{\beta_t^2}\langle A_0, A_t\rangle)}]
\end{equation}

With taylor expansion, 
\begin{equation}
\mathbb{E}_{A_0\sim p_0}[A_0\exp{(\frac{\alpha_t}{\beta_t^2}\langle A_0, A_t\rangle)}]
=\sum_{k=0}^\infty\frac{\alpha_t^k}{\beta_t^{2k}k!} \mathbb{E}_{A_0\sim p_0}[A_0\langle A_0, A_t\rangle^k]
\end{equation}

Let $F_k(A_t)=\mathbb{E}_{A_0\sim p_0}[A_0\langle A_0, A_t\rangle^k]$, then
\begin{align}
F_k(A_t)_{ij}=\sum_{\mathbf{a}\in [n]^{2k}}\mathbb{E}_{A_0\sim p_0} {A_0}_{ij}\prod_{l=1}^k {A_0}_{a_{2l-1}a_{2l}}\prod_{l=1}^k {A_t}_{a_{2l-1}a_{2l}}
\end{align}
As $p_0$ is permutation-invariant, we can replace $A_0$ with $\pi^{-1}(A_0)$, then,
\begin{align}
F_k(A_t)_{ij}=\sum_{\mathbf{a}\in [n]^{2k}}\prod_{l=1}^k {A_t}_{a_{2l-1}a_{2l}} \frac{1}{n!}\\
\sum_{\pi\in \Pi_n}\mathbb{E}_{A_0\sim p_0} {A_0}_{\pi(i)\pi(j)}    \prod_{l=1}^k {A_0}_{\pi(a_{2l-1})\pi(a_{2l})}\\
=\sum_{\mathbf{a}\in [n]^{2k}}\prod_{l=1}^k {A_t}_{a_{2l-1}a_{2l}} \mathbb{E}_{A_0\sim p_0}Q(A_0,S_{ij\mathbf{a}}),
\end{align}
where $Q$ is equivariant graph polynomial basis, $S_{ij\mathbf{a}}$ is a graph with edge $(i,j)$ and $(a_{2l-1}, a_{2l})$, where parallel edge is allowed. However, as elements in $A_0$ are $0, 1$, $Q(A_0, S_{ij\mathbf{a}})=Q(A_0, \bar S_{ij\mathbf{a}})$, where $\bar S_{ij\mathbf{a}}$ is $S_{ij\mathbf{a}}$ without parallel edges.

Therefore,
\begin{align}
    F_k(A_t) = \sum_{ij\in [n]^2}E_{ij}\sum_{\mathbf{a}\in [n]^{2k}}\prod_{l=1}^k {A_t}_{a_{2l-1}a_{2l}} \mathbb{E}_{A_0\sim p_0}Q(A_0,S_{ij\mathbf{a}}),
\end{align}
where $E_{ij}$ is a $n\times n$ matrix with its $(i,j)$ element being $1$ and all other elements are $0$. 

As $p_0$ is permutation-invariant, $p_t$ is also permutation-equivariant. Therefore, $F_k(A_t)$ is permutation-equivariant
\begin{align}
    F_k(A_t) &= \frac{1}{n!}\sum_{\pi\in \Pi_n}\pi^{-1}(F_k(\pi(A_t)))\\
    F_k(A_t)_{ij} &= \frac{1}{n!}\sum_{\pi\in \Pi_n}F_k(\pi(A_t))_{\pi(i)\pi(j)}\\
    &=\frac{1}{n!}\sum_{\pi\in \Pi_n}\sum_{\mathbf{a}\in [n]^{2k}}\prod_{l=1}^k {A_t}_{\pi^{-1}(a_{2l-1})\pi^{-1}(a_{2l})} \mathbb{E}_{A_0\sim p_0}Q(A_0,S_{\pi(i)\pi(j)\mathbf{a}})
\end{align}
Note that $S_{\pi(i)\pi(j)\mathbf{a}}$ is isomorphic to $S_{ij\pi^{-1}(\mathbf{a})}$. So
\begin{align}
F_k(A_t)_{ij} &=\frac{1}{n!}\sum_{\pi\in \Pi_n}\sum_{\mathbf{a}\in [n]^{2k}}\prod_{l=1}^k {A_t}_{\pi^{-1}(a_{2l-1})\pi^{-1}(a_{2l})} \mathbb{E}_{A_0\sim p_0}Q(A_0,S_{ij\pi^{-1}(\mathbf{a}}))
\end{align}
Replace $\pi^{-1}$ with $\pi$,
\begin{align}
F_k(A_t) &=\sum_{ij\in [n]^2}E_{ij}\frac{1}{n!}\sum_{\pi\in \Pi_n}\sum_{\mathbf{a}\in [n]^{2k}}\prod_{l=1}^k {A_t}_{\pi(a_{2l-1})\pi(a_{2l})} \mathbb{E}_{A_0\sim p_0}Q(A_0,S_{ij\pi(\mathbf{a})}),\\
&= \frac{1}{n!}\sum_{\pi\in \Pi_n}\sum_{\mathbf{a}\in [n]^{2k}}\sum_{ij\in [n]^2}\prod_{l=1}^k {A_t}_{\pi(a_{2l-1})\pi(a_{2l})}E_{ij} \mathbb{E}_{A_0\sim p_0}Q(A_0,S_{ij\pi(\mathbf{a})}),\\
&= \frac{1}{n!}\sum_{\pi\in \Pi_n}\sum_{\mathbf{a}\in [n]^{2k}}\sum_{ij\in [n]^2}\prod_{l=1}^k {A_t}_{\pi(a_{2l-1})\pi(a_{2l})}E_{\pi(i)\pi(j)} \mathbb{E}_{A_0\sim p_0}Q(A_0,S_{\pi(i)\pi(j)\pi(\mathbf{a})}),\\
\end{align}
Note that $S_{\pi(i)\pi(j)\pi(\mathbf{a})}$ is isomorphic to $S_{ij\mathbf{a}}$.
\begin{align}
F_k(A_t) &= \sum_{ij\in [n]^2}\sum_{\mathbf{a}\in [n]^{2k}}\mathbb{E}_{A_0\sim p_0}Q(A_0,S_{ij\mathbf{a}})\frac{1}{n!}\sum_{\pi\in \Pi_n}\prod_{l=1}^k {A_t}_{\pi(a_{2l-1})\pi(a_{2l})}E_{\pi(i)\pi(j)},\\
&= \sum_{ij\in [n]^2}\sum_{\mathbf{a}\in [n]^{2k}}\mathbb{E}_{A_0\sim p_0}Q(A_0,S_{ij\mathbf{a}})\tilde Q(A_t, T_{ij\mathbf{a}}),\\
\end{align}
where $\tilde Q$ is equivariant graph polynomial basis, and $T_{ij\mathbf{a}}$ is a graph with red node $i, j$ and edge $\{(a_{2l-1}, a_{2l})|l=1,2,...,k\}$.

Putting it altogether,
\begin{align}
&\mathbb{E}_{A_0\sim p_0}p_t(A_t|A_0)A_0 
= \frac{\exp{(-\frac{m\alpha_t^2}{2\beta_t^2})}}{\beta_t\sqrt{2\pi}}\exp{(-\frac{\Vert A_t\Vert_F^2}{2\beta_t^2})}\\
&
\sum_{k=0}^\infty\sum_{ij\in [n]^2}\sum_{\mathbf{a}\in [n]^{2k}}\frac{\alpha_t^k}{\beta_t^{2k}k!} [\mathbb{E}_{A_0\sim p_0}Q(A_0,S_{ij\mathbf{a}})]\tilde Q(A_t, T_{ij\mathbf{a}})
\end{align}

Similarly,

\begin{align}
&\mathbb{E}_{A_0\sim p_0}p_t(A_t|A_0)
= \frac{\exp{(-\frac{m\alpha_t^2}{2\beta_t^2})}}{\beta_t\sqrt{2\pi}}\exp{(-\frac{\Vert A_t\Vert_F^2}{2\beta_t^2})}\\
&
\sum_{k=0}^\infty\sum_{ij\in [n]^2}\sum_{\mathbf{a}\in [n]^{2k}}\frac{\alpha_t^k}{\beta_t^{2k}k!} [\mathbb{E}_{A_0\sim p_0}Q(A_0,S_{\mathbf{a}})]Q(A_t, S_{\mathbf{a}})
\end{align}


\end{document}